\documentclass[11pt,a4paper]{article}
\usepackage[hyperref]{naaclhlt2019}
\usepackage{times}
\usepackage{latexsym}

\hypersetup{draft}

\aclfinalcopy 
\def\aclpaperid{830} 

\usepackage{graphicx}
\usepackage{subfig}
\usepackage{bm}
\usepackage{amssymb}
\usepackage{amsmath}
\DeclareMathOperator*{\onehot}{one-hot}
\DeclareMathOperator*{\argmax}{arg\,max}
\DeclareMathOperator*{\softmax}{softmax}
\DeclareMathOperator*{\uniform}{Uniform}
\DeclareMathOperator*{\gumbel}{Gumbel}

\newcommand{\vc}{\bm{c}}
\newcommand{\ve}{\bm{e}}
\newcommand{\vf}{\bm{f}}
\newcommand{\vh}{\bm{h}}

\newcommand{\vE}{\bm{E}}

\newcommand{\vtheta}{\bm{\theta}}
\newcommand{\loss}{\mathcal{L}}
\newcommand{\bftab}{\fontseries{b}\selectfont}

\title{Bi-Directional Differentiable Input Reconstruction \\ for Low-Resource Neural Machine Translation}

\author{Xing Niu \\
	University of Maryland \\
	{\tt \href{mailto:xingniu@cs.umd.edu}{xingniu@cs.umd.edu}} \\\And
	Weijia Xu \\
	University of Maryland \\
	{\tt \href{mailto:weijia@cs.umd.edu}{weijia@cs.umd.edu}} \\\And
	Marine Carpuat \\
	University of Maryland \\
	{\tt \href{mailto:marine@cs.umd.edu}{marine@cs.umd.edu}} \\}

\date{}

\begin{document}
\maketitle

\begin{abstract}
We aim to better exploit the limited amounts of parallel text available in low-resource settings by introducing a differentiable reconstruction loss for neural machine translation (NMT). This loss compares original inputs to reconstructed inputs, obtained by back-translating translation hypotheses into the input language. We leverage differentiable sampling and bi-directional NMT to train models end-to-end, without introducing additional parameters. This approach achieves small but consistent BLEU improvements on four language pairs in both translation directions, and outperforms an alternative differentiable reconstruction strategy based on hidden states.
\end{abstract}

\section{Introduction}

Neural Machine Translation (NMT) performance degrades sharply when parallel training data is limited \citep{KoehnK17}. Past work has addressed this problem by leveraging monolingual data \citep{SennrichHB16,RamachandranLL17} or multilingual parallel data \cite{ZophYMK16,JohnsonSLKWCTVW17,GuHDL18}. We hypothesize that the traditional training can be complemented by better leveraging limited training data. To this end, we propose a new training objective for this model by augmenting the standard translation cross-entropy loss with a \textbf{differentiable input reconstruction loss} to further exploit the source side of parallel samples.

Input reconstruction is motivated by the idea of round-trip translation. Suppose sentence $\vf$ is translated forward to $\ve$ using model $\vtheta_{fe}$ and then translated back to $\hat{\vf}$ using model $\vtheta_{ef}$, then $\ve$ is more likely to be a good translation if the distance between $\hat{\vf}$ and $\vf$ is small \citep{Brislin70}. Prior work applied round-trip translation to monolingual examples and sampled the intermediate translation $\ve$ from a $K$-best list generated by model $\vtheta_{fe}$ using beam search \citep{ChengXHHWSL16,HeXQWYLM16}. However, beam search is not differentiable which prevents back-propagating reconstruction errors to $\vtheta_{fe}$. As a result, reinforcement learning algorithms, or independent updates to $\vtheta_{fe}$ and $\vtheta_{ef}$ were required.

In this paper, we focus on the problem of making input reconstruction differentiable to simplify training. In past work, \citet{TuLSLL17} addressed this issue by reconstructing source sentences from the decoder's hidden states. However, this reconstruction task can be artificially easy if hidden states over-memorize the input. This approach also requires a separate auxiliary reconstructor, which introduces additional parameters.

We propose instead to combine benefits from differentiable sampling and bi-directional NMT to obtain a compact model that can be trained end-to-end with back-propagation. Specifically,
\begin{itemize}
	\item Translations are sampled using the Straight-Through Gumbel Softmax (STGS) estimator \citep{JangGP17,BengioLC13}, which allows back-propagating reconstruction errors.
	\item Our approach builds on the bi-directional NMT model \citep{NiuDC18,JohnsonSLKWCTVW17}, which improves low-resource translation by jointly modeling translation in both directions (e.g., Swahili $\leftrightarrow$ English). A single bi-directional model is used as a translator and a reconstructor (i.e. $\vtheta_{ef}=\vtheta_{fe}$) without introducing more parameters. 
\end{itemize}

Experiments show that our approach outperforms reconstruction from hidden states. It achieves consistent improvements across various low-resource language pairs and directions, showing its effectiveness in making better use of limited parallel data.

\section{Background}

Using round-trip translations ($\vf$$\,\rightarrow\,$$\ve$$\,\rightarrow\,$$\hat{\vf}$) as a training signal for NMT usually requires auxiliary models to perform back-translation and cannot be trained end-to-end without reinforcement learning. For instance, \citet{ChengXHHWSL16} added a reconstruction loss for monolingual examples to the training objective. \citet{HeXQWYLM16} evaluated the quality of $\ve$ by a language model and $\hat{\vf}$ by a reconstruction likelihood. Both approaches have symmetric forward and backward translation models which are updated alternatively. This require policy gradient algorithms for training, which are not always stable.

Back-translation \citep{SennrichHB16} performs half of the reconstruction process, by generating a synthetic source side for monolingual target language examples: $\ve\,\rightarrow\,\hat{\vf}$. It uses an auxiliary backward model to generate the synthetic data but only updates the parameters of the primary forward model. Iteratively updating forward and backward models \citep{ZhangLZC18,NiuDC18} is an expensive solution as back-translations are regenerated at each iteration.

Prior work has sought to simplify the optimization of reconstruction losses by side-stepping beam search. \citet{TuLSLL17} first proposed to reconstruct NMT input from the decoder's hidden states while \citet{WangTSZGL18, WangTWL18} suggested to use both encoder and decoder hidden states to improve translation of dropped pronouns. However, these models might achieve low reconstruction errors by learning to copy the input to hidden states. To avoid copying the input, \citet{ArtetxeLAC18} and \citet{LampleDR18} used denoising autoencoders \citep{VincentLBM08} in unsupervised NMT.

Our approach is based instead on the Gumbel Softmax \citep{JangGP17,MaddisonMT17}, which facilitates differentiable sampling of sequences of discrete tokens. It has been successfully applied in many sequence generation tasks, including artificial language emergence for multi-agent communication \citep{HavrylovT17}, composing tree structures from text \citep{ChoiYL18}, and tasks under the umbrella of generative adversarial networks \cite{GoodfellowPMXWOCB14} such as generating the context-free grammar \citep{KusnerH16}, machine comprehension \citep{WangLZ17} and machine translation \citep{GuIL18}.

\section{Approach}

NMT is framed as a conditional language model, where the probability of predicting target token $e_t$ at step $t$ is conditioned on the previously generated sequence of tokens $\ve_{<t}$ and the source sequence $\vf$ given the model parameter $\vtheta$. Suppose each token is indexed and represented as a one-hot vector, its probability is realized as a softmax function over a linear transformation $a(\vh_t)$ where $\vh_t$ is the decoder's hidden state at step $t$:
\begin{equation}\label{eq:token_prob}
P(e_t|\ve_{<t},\vf;\vtheta)=\softmax(a(\vh_t))^\top e_t.
\end{equation}
The hidden state is calculated by a neural network $g$ given the embeddings of the previous target tokens $\ve_{<t}$ in the embedding matrix $\vE(\ve_{<t})$ and the context $\vc_t$ coming from the source:
\begin{equation}\label{eq:hidden_fuction}
\vh_t=g(\vE(\ve_{<t}),\vc_t).
\end{equation}

In our bi-directional model, the source sentence can be either $\vf$ or $\ve$ and is respectively translated to $\ve$ or $\vf$. The language is marked by a tag (e.g., \texttt{<en>}) at the beginning of each source sentence \citep{JohnsonSLKWCTVW17,NiuDC18}. To facilitate symmetric reconstruction, we also add language tags to target sentences. The training data corpus is then built by swapping the source and target sentences of a parallel corpus and appending the swapped version to the original.

\subsection{Bi-Directional Reconstruction}

Our bi-directional model performs both forward translation and backward reconstruction. By contrast, uni-directional models require an auxiliary reconstruction module, which introduces additional parameters. This module can be either a decoder-based reconstructor \citep{TuLSLL17,WangTSZGL18,WangTWL18} or a reversed dual NMT model \citep{ChengXHHWSL16,HeXQWYLM16,WangXZBQLL18,ZhangLZC18}.

Here the reconstructor, which shares the same parameter with the translator $T(\cdot)$, can also be trained end-to-end by maximizing the log-likelihood of reconstructing $\vf$:
\begin{equation}\label{eq:reconstruction_obj}
\loss_R = \sum_{\vf} \log P(\vf\,|\,T(\vf;\vtheta);\vtheta),
\end{equation}
Combining with the forward translation likelihood
\begin{equation}\label{eq:translation_obj}
\loss_T = \sum_{(\vf\parallel\ve)} \log P(\ve\,|\,\vf;\vtheta),
\end{equation}
we use $\loss=\loss_T+\loss_R$ as the final training objective for $\vf\rightarrow\ve$. The dual $\ve\rightarrow\vf$ model is trained simultaneously by swapping the language direction in bi-directional NMT.

Reconstruction is reliable only with a model that produces reasonable base translations. Following prior work \citep{TuLSLL17,HeXQWYLM16,ChengXHHWSL16}, we pre-train a base model with $\loss_T$ and fine-tune it with $\loss_T+\loss_R$.

\subsection{Differentiable Sampling}
\label{sec:sampling}

We use differentiable sampling to side-step beam search and back-propagate error signals. We use the Gumbel-Max reparameterization trick \citep{MaddisonTM14} to sample a translation token at each time step from the softmax distribution in Equation \ref{eq:token_prob}:
\begin{equation}\label{eq:gumbel_max}
e_t=\onehot\Big(\argmax_k\big(a(\vh_t)_k+G_k\big)\Big)
\end{equation}
where $G_k$ is i.i.d. and drawn from $\gumbel(0,1)$\footnote{i.e. $G_k=-\log(-\log(u_k))$ and $u_k\sim \uniform(0,1)$.}. We use scaled $\gumbel$ with parameter $\beta$, i.e. $\gumbel(0,\beta)$, to control the randomness. The sampling becomes deterministic (which is equivalent to greedy search) as $\beta$ approaches 0.

Since $\argmax$ is not a differentiable operation, we approximate its gradient with the Straight-Through Gumbel Softmax (STGS) \citep{JangGP17,BengioLC13}: $\nabla_{\vtheta}e_t \approx \nabla_{\vtheta}\tilde{e}_t$, where
\begin{equation}\label{eq:gumbel_softmax}
\tilde{e}_t=\softmax\big((a(\vh_t)+G)/\tau\big)
\end{equation}
As $\tau$ approaches 0, $\softmax$ is closer to $\argmax$ but training might be more unstable. While the STGS estimator is biased when $\tau$ is large,
it performs well in practice \citep{GuIL18,ChoiYL18} and is sometimes faster and more effective than reinforcement learning \citep{HavrylovT17}.

To generate coherent intermediate translations, the decoder used for sampling only consumes its previously predicted $\hat{\ve}_{<t}$. This contrasts with the usual \textit{teacher forcing} strategy \citep{WilliamsZ89}, which always feeds in the ground-truth previous tokens $\ve_{<t}$ when predicting the current token $\hat{e}_t$. With teacher forcing, the sequence concatenation $[\ve_{<t};\hat{e}_t]$ is probably coherent at each time step, but the actual predicted sequence $[\hat{\ve}_{<t};\hat{e}_t]$ would break the continuity.\footnote{Sampling with teacher forcing yielded consistently worse BLEU than baselines in preliminary experiments.}

\section{Experiments}

\begin{table}[t]
	\begin{center}
		\begin{tabular}{l|r|r|r}
		\# sent. & Training & Dev. & Test \\
			\hline
			\texttt{SW$\leftrightarrow$EN} & 60,570 & 500 & 3,000 \\
			\texttt{TL$\leftrightarrow$EN} & 70,703 & 704 & 3,000 \\
			\texttt{SO$\leftrightarrow$EN} & 68,550 & 844 & 3,000 \\
			\texttt{TR$\leftrightarrow$EN} & 207,021 & 1,001 & 3,007 \\
		\end{tabular}
	\end{center}
	\caption{Experiments are conducted on four low-resource language pairs, in both translation directions.}
	\label{tab:data}
\end{table}

\begin{table*}[t]
	\begin{center}
		\scalebox{0.75}{
			\begin{tabular}{l|rr|rr|rr|rr}
				Model & \texttt{EN}$\rightarrow$\texttt{SW} & \texttt{SW}$\rightarrow$\texttt{EN} & \texttt{EN}$\rightarrow$\texttt{TL} & \texttt{TL}$\rightarrow$\texttt{EN} & \texttt{EN}$\rightarrow$\texttt{SO} & \texttt{SO}$\rightarrow$\texttt{EN} & \texttt{EN}$\rightarrow$\texttt{TR} & \texttt{TR}$\rightarrow$\texttt{EN} \\
				\hline
				Baseline & 33.60 $\pm$ 0.14 & 30.70 $\pm$ 0.19 & 27.23 $\pm$ 0.11 & 32.15 $\pm$ 0.21 & 12.25 $\pm$ 0.08 & 20.80 $\pm$ 0.12 & 12.90 $\pm$ 0.04 & 15.32 $\pm$ 0.11 \\
				\hline
				\textsc{Hidden} & 33.41 $\pm$ 0.15 & 30.91 $\pm$ 0.19 & 27.43 $\pm$ 0.14 & 32.20 $\pm$ 0.35 & 12.30 $\pm$ 0.11 & 20.72 $\pm$ 0.16 & 12.77 $\pm$ 0.11 & 15.34 $\pm$ 0.10 \\
				\hspace{10pt}$\Delta$ & \color{red} -0.19 $\pm$ 0.24 & \bftab \color{blue} 0.21 $\pm$ 0.14 & \bftab \color{blue} 0.19 $\pm$ 0.13 & 0.04 $\pm$ 0.17 & 0.05 $\pm$ 0.11 & \color{red} -0.08 $\pm$ 0.12 & \color{red} -0.13 $\pm$ 0.13 & 0.01 $\pm$ 0.07 \\
				\hline
				$\beta=0$ & 33.92 $\pm$ 0.10 & 31.37 $\pm$ 0.18 & 27.65 $\pm$ 0.09 & 32.75 $\pm$ 0.32 & 12.47 $\pm$ 0.08 & 21.14 $\pm$ 0.19 & 13.26 $\pm$ 0.07 & 15.60 $\pm$ 0.19 \\
				\hspace{10pt}$\Delta$ & \bftab \color{blue} 0.32 $\pm$ 0.12 & \bftab \color{blue} 0.66 $\pm$ 0.11 & \bftab \color{blue} 0.42 $\pm$ 0.16 & \bftab \color{blue} 0.59 $\pm$ 0.13 & \bftab \color{blue} 0.22 $\pm$ 0.04 & \bftab \color{blue} 0.35 $\pm$ 0.15 & \bftab \color{blue} 0.36 $\pm$ 0.09 & \bftab \color{blue} 0.28 $\pm$ 0.11 \\
				\hline
				$\beta=0.5$ & 33.97 $\pm$ 0.08 & 31.39 $\pm$ 0.09 & 27.65 $\pm$ 0.10 & 32.65 $\pm$ 0.24 & 12.48 $\pm$ 0.09 & 21.20 $\pm$ 0.14 & 13.16 $\pm$ 0.08 & 15.52 $\pm$ 0.07 \\
				\hspace{10pt}$\Delta$ & \bftab \color{blue} 0.37 $\pm$ 0.09 & \bftab \color{blue} 0.69 $\pm$ 0.11 & \bftab \color{blue} 0.42 $\pm$ 0.11 & \bftab \color{blue} 0.50 $\pm$ 0.08 & \bftab \color{blue} 0.23 $\pm$ 0.03 & \bftab \color{blue} 0.41 $\pm$ 0.13 & \bftab \color{blue} 0.25 $\pm$ 0.09 & \bftab \color{blue} 0.19 $\pm$ 0.05 \\
		\end{tabular}}
	\end{center}
	\caption{BLEU scores on eight translation directions. The numbers before and after `$\pm$' are the mean and standard deviation over five randomly seeded models. Our proposed methods ($\beta=0/0.5$) achieve small but consistent improvements. $\Delta$BLEU scores are in  bold if mean$-$std is above zero while in red if the mean is below zero.}
	\label{tab:bleu}
\end{table*}

\subsection{Tasks and Data}

We evaluate our approach on four low-resource language pairs. Parallel data for Swahili$\leftrightarrow$English (\texttt{SW$\leftrightarrow$EN}), Tagalog$\leftrightarrow$English (\texttt{TL$\leftrightarrow$EN}) and Somali$\leftrightarrow$English (\texttt{SO$\leftrightarrow$EN}) contains a mixture of domains such as news and weblogs and is collected from the IARPA MATERIAL program\footnote{\url{https://www.iarpa.gov/index.php/research-programs/material}}, the Global Voices parallel corpus\footnote{\url{http://casmacat.eu/corpus/global-voices.html}}, Common Crawl \citep{SmithSPKCL13}, and the LORELEI Somali representative language pack (LDC2018T11). The test samples are extracted from the held-out ANALYSIS set of MATERIAL. Parallel Turkish$\leftrightarrow$English (\texttt{TR$\leftrightarrow$EN}) data is provided by the WMT news translation task \citep{BojarFFGHHKM18}. We use pre-processed ``corpus'', ``newsdev2016'', ``newstest2017'' as training,  development and test sets.\footnote{\url{http://data.statmt.org/wmt18/translation-task/preprocessed/}}

We apply normalization, tokenization, true-casing, joint source-target BPE with 32,000 operations \citep{SennrichHB16a} and sentence-filtering (length 80 cutoff) to parallel data.
Itemized data statistics after preprocessing can be found in Table~\ref{tab:data}. We report case-insensitive BLEU with the WMT standard `13a' tokenization using SacreBLEU \citep{Post18}.

\subsection{Model Configuration and Baseline}

We build NMT models upon the attentional RNN encoder-decoder architecture \citep{BahdanauCB15} implemented in the Sockeye toolkit \citep{CoRR:Sockeye}. Our translation model uses a bi-directional encoder with a single LSTM layer of size 512, multilayer perceptron attention with a layer size of 512, and word representations of size 512. We apply layer normalization \citep{BaKH16} and add dropout to embeddings and RNNs \citep{GalG16} with probability 0.2. We train using the Adam optimizer \citep{KingmaB15} with a batch size of 48 sentences and we checkpoint the model every 1000 updates. 
The learning rate for baseline models is initialized to 0.001 and reduced by 30\% after 4 checkpoints without improvement of perplexity on the development set. Training stops after 10 checkpoints without improvement.

The bi-directional NMT model ties source and target embeddings to yield a bilingual vector space. It also ties the output layer's weights and embeddings to achieve better performance in low-resource scenarios \citep{PressW17,NguyenC18}.

We train five randomly seeded bi-directional baseline models by optimizing the forward translation objective $\loss_T$ and report the mean and standard deviation of test BLEU. We fine-tune baseline models with objective $\loss_T+\loss_R$, inheriting all settings except the learning rate which is re-initialized to 0.0001. Each randomly seeded model is fine-tuned independently, so we are able to report the standard deviation of $\Delta$BLEU.

\subsection{Contrastive Reconstruction Model}

We compare our approach with reconstruction from hidden states (\textsc{Hidden}). Following the best practice of \citet{WangTSZGL18}, two reconstructors are used to take hidden states from both the encoder and the decoder. The corresponding two reconstruction losses and the canonical translation loss were originally uniformly weighted (i.e. $1,1,1$), but we found that balancing the reconstruction and translation losses yields better results (i.e. $0.5,0.5,1$) in preliminary experiments.\footnote{We observed around 0.2 BLEU gains for \texttt{TR$\leftrightarrow$EN} tasks.}

We use the reconstructor exclusively to compute the reconstruction training loss. It has also been used to re-rank translation hypotheses in prior work, but \citet{TuLSLL17} showed in ablation studies that the gains from re-ranking are small compared to those from training.

\begin{figure*}[t]
	\centering
	\subfloat[training set]{
		\label{fig:ppl-train}
		\includegraphics[width=0.49\textwidth]{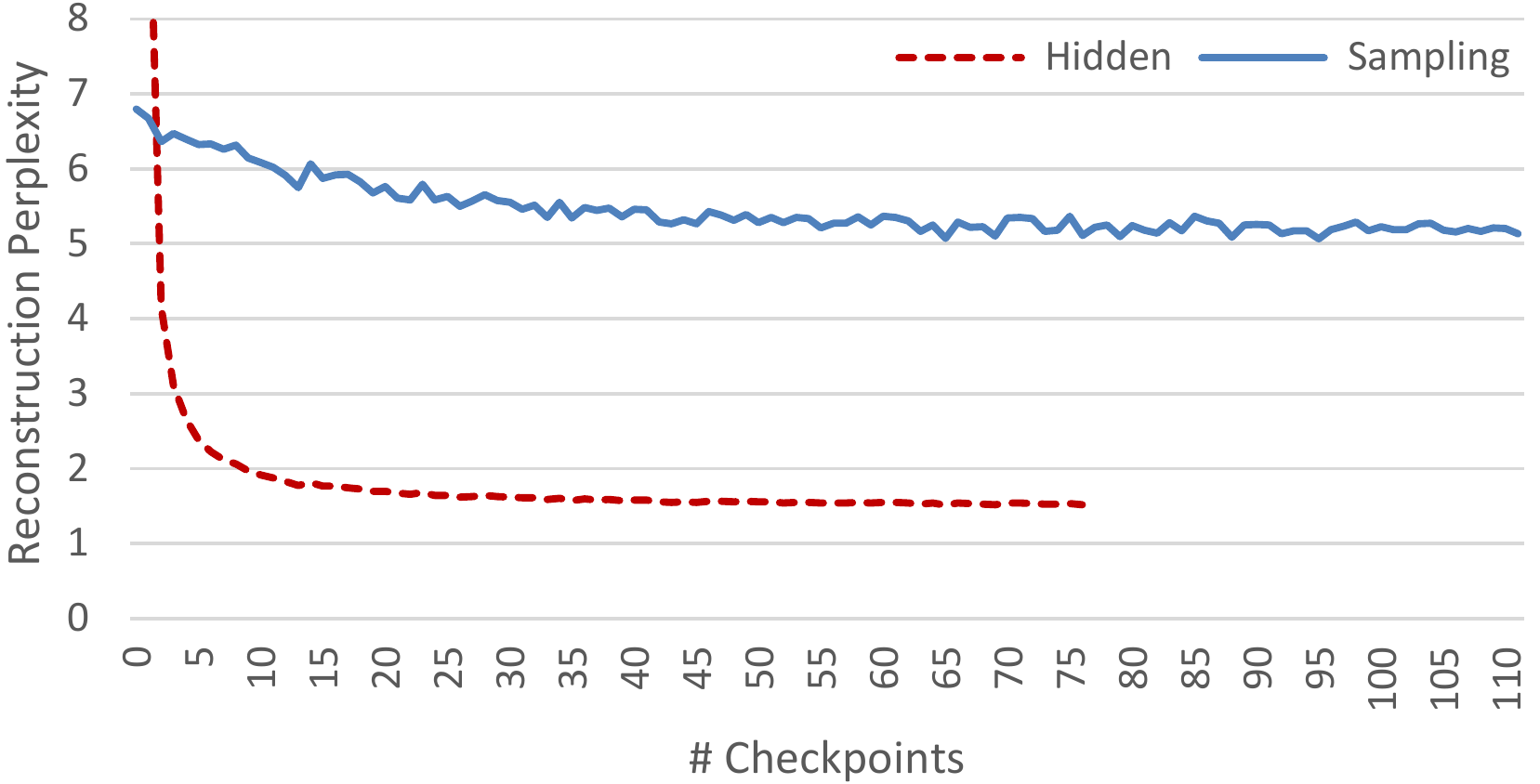}
	}
	\subfloat[development set]{
		\label{fig:ppl-dev}
		\includegraphics[width=0.49\textwidth]{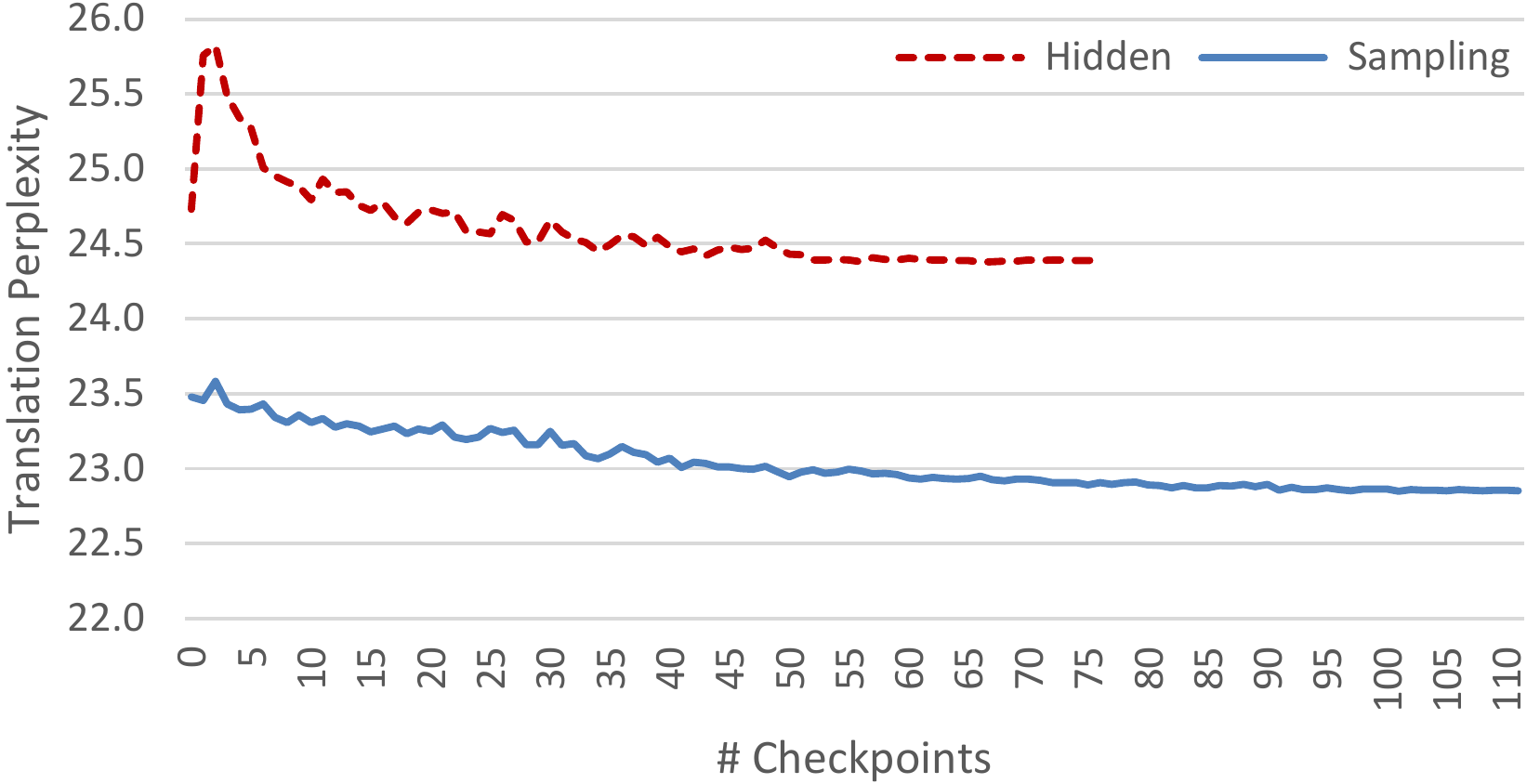}
	}
	\caption{Training curves of perplexity on the training and the development sets for \texttt{TR$\leftrightarrow$EN}. Reconstructing from hidden states (\textsc{Hidden}) and reconstructing from sampled translations ($\beta=0$) are compared. \textsc{Hidden} achieves extremely low training perplexity and suffers from unstable training during the early stage.}
	\label{fig:ppl}
\end{figure*}

\subsection{Results}

Table \ref{tab:bleu} shows that our reconstruction approach achieves small but consistent BLEU improvements over the baseline on all eight tasks.\footnote{The improvements are significant with $p<0.01$.}

We evaluate the impact of the Gumbel Softmax hyperparameters on the development set. We select $\tau=2$ and $\beta=0/0.5$ based on training stability and BLEU. Greedy search (i.e. $\beta=0$) performs similarly as sampling with increased Gumbel noise (i.e. more random translation selection when $\beta=0.5$): increased randomness in sampling does not have a strong impact on BLEU, even though random sampling may approximate the data distribution better \cite{OttAGR18}. We hypothesize that more random translation selection introduces lower quality samples and therefore noisier training signals. This is consistent with the observation that random sampling is less effective for back-translation in low-resource settings \cite{EdunovOAG18}.

Sampling-based reconstruction is effective even if there is moderate domain mismatch between the training and the test data, such as in the case that the word type out-of-vocabulary (OOV) rate of \texttt{TR}$\rightarrow$\texttt{EN} is larger than 20\%. Larger improvements can be achieved when the test data is closer to training examples. For example, the OOV rate of \texttt{SW}$\rightarrow$\texttt{EN} is much smaller than the OOV rate of \texttt{TR}$\rightarrow$\texttt{EN} and the former obtains higher $\Delta$BLEU.

Our approach yields more consistent results than reconstructing from hidden states. The latter fails to improve BLEU in more difficult cases, such as \texttt{TR}$\leftrightarrow$\texttt{EN} with high OOV rates. We observe extremely low training perplexity for \textsc{Hidden} compared with our proposed approach (Figure \ref{fig:ppl-train}). This suggests that \textsc{Hidden} yields representations that memorize the input rather than improve output representations.

Another advantage of our approach is that all parameters were jointly pre-trained, which results in more stable training behavior. By contrast, reconstructing from hidden states requires to initialize the reconstructors independently and suffers from unstable early training behavior (Figure \ref{fig:ppl}).

\section{Conclusion}

We studied reconstructing the input of NMT from its intermediate translations to better exploit training samples in low-resource settings. We used a bi-directional NMT model and the Straight-Through Gumbel Softmax to build a fully differentiable reconstruction model that does not require any additional parameters. We empirically demonstrated that our approach is effective in low-resource scenarios. In future work, we will investigate the use of differentiable reconstruction from sampled sequences in unsupervised and semi-supervised sequence generation tasks. In particular, we will exploit monolingual corpora in addition to parallel corpora for NMT.

\section*{Acknowledgments}

We thank the three anonymous reviewers for their helpful comments and suggestions. We also thank the members of the Computational Linguistics and Information Processing (CLIP) lab at the University of Maryland for helpful discussions.

This research is based upon work supported in part by an Amazon Web Services Machine Learning Research Award, and by the Office of the Director of National Intelligence (ODNI), Intelligence Advanced Research Projects Activity (IARPA), via contract \#FA8650-17-C-9117. The views and conclusions contained herein are those of the authors and should not be interpreted as necessarily representing the official policies, either expressed or implied, of ODNI, IARPA, or the U.S. Government. The U.S. Government is authorized to reproduce and distribute reprints for governmental purposes notwithstanding any copyright annotation therein.

\bibliography{naaclhlt2019}
\bibliographystyle{acl_natbib}

\end{document}